\documentclass[10pt, a4paper]{article}
\usepackage{lrec2022} 
\usepackage{nert}
\usepackage{multibib}
\newcites{languageresource}{Language Resources}
\usepackage{graphicx}
\usepackage{natbib}
\usepackage{tabularx}
\usepackage{soul}
\usepackage{titlesec}
\titleformat{\subsection}{\normalfont\SmallTitleFont\bf\raggedright}{\thesubsection.}{1em}{}
\titleformat{\subsubsection}{\normalfont\normalsize\bf\raggedright}{\thesubsubsection.}{1em}{}
\renewcommand\thesection{\arabic{section}}
\renewcommand\thesubsection{\thesection.\arabic{subsection}}
\renewcommand\thesubsubsection{\thesubsection.\arabic{subsubsection}}

\usepackage{epstopdf}

\usepackage{hyperref}
\usepackage{xstring}

\usepackage{color}

\newcommand{\amr}[1]{\texttt{#1}}

\title{Spanish Abstract Meaning Representation: \\ Annotation of a General Corpus}

\name{\begin{tabular}{c}Shira Wein\textsuperscript{1}, Lucia Donatelli\textsuperscript{2}, Ethan Ricker\textsuperscript{1},\\Calvin Engstrom\textsuperscript{1}, Alex Nelson\textsuperscript{1}, Nathan Schneider\textsuperscript{1}\end{tabular}}

\address{
\textsuperscript{1}Georgetown University, USA \\
\{\emldisplay{sw1158@georgetown.edu}{sw1158}, \emldisplay{ear131@georgetown.edu}{ear131}, \emldisplay{cle41@georgetown.edu}{cle41}, \emldisplay{amn106@georgetown.edu}{amn106}, \emldisplay{nathan.schneider@georgetown.edu}{nathan.schneider}\}\texttt{@georgetown.edu}\\
\textsuperscript{2}Saarland University, Germany \\
\eml{donatelli@coli.uni-saarland.de}\\}

\abstract{
The Abstract Meaning Representation (AMR) formalism, designed originally for English, has been adapted to a number of languages. We build on previous work proposing the annotation of AMR in Spanish, which resulted in the release of 50 Spanish AMR annotations for the fictional text \emph{The Little Prince}. In this work, we present the first sizable, general annotation project for Spanish Abstract Meaning Representation. Our approach to annotation makes use of Spanish rolesets from the AnCora-Net lexicon and extends English AMR with semantic features specific to Spanish. In addition to our guidelines, we release an annotated corpus (586 annotations total, for 486 unique sentences) of multiple genres of documents from the ``Abstract Meaning Representation 2.0 - Four Translations'' sembank. This corpus is commonly used for evaluation of AMR parsing and generation, but does not include gold AMRs; we hope that providing gold annotations for this dataset can result in a more complete approach to cross-lingual AMR parsing.  
Finally, we perform a disagreement analysis and discuss the implications of our work on the adaptability of AMR to languages other than English.
\\ \newline  }

\begin{document}

\maketitleabstract

\section{Introduction}




The Abstract Meaning Representation (AMR) semantic formalism represents the core meaning of a sentence as a directed, rooted graph \citep{banarescu-etal-2013-abstract}. While there are large AMR-annotated corpora available for English, if AMR is to be useful as an interlingua \citep{xue-etal-2014-interlingua}, cross-lingual adaptations of AMR are necessary to effectively capture meaning of other languages with the AMR structure. This includes evaluating cross-lingual efficacy of rolesets, word senses, and whether or not AMR core and non-core relations can effectively capture ``who is doing what to whom'' when construed differently than in English.

Recent work has adapted AMR to a variety of languages (\cref{ssec:adaptations}).
Though Spanish is one of the most spoken languages in the world, there has only been one previous proposal for adapting it to AMR: \Citet{migueles-abraira-etal-2018-annotating} presented a corpus of 50 annotations 
focusing on a set of linguistic phenomena that need to be added to English AMR to provide coverage for Spanish semantic phenomena. We extend this prior work on Spanish AMR and present the first substantial Spanish AMR corpus with more than 500 annotations. Our design goals are detailed in the accompanying annotation scheme that accounts for a range of semantically-meaningful linguistic phenomena in Spanish beyond what was previously proposed. 

Though many AMR corpora (for English as well as for other languages) are on sentences from translations of Saint-Exup\'{e}ry's \emph{Le Petit Prince} (\emph{The Little Prince}), this is a work of fiction with a very specific writing style and limited expressivity. We thus explore annotation for more general-purpose corpora. Specifically, we annotate the Spanish sentences from the ``Abstract Meaning Representation 2.0 - Four Translations'' dataset \citep{damonte_cohen_amr_four_translations}, a corpus from the news domain that 
has become a popular resource for evaluation of cross-lingual AMR parsers \citep{blloshmi-etal-2020-xl,procopio-etal-2021-sgl,cai-etal-2021-multilingual-amr}. Three trained annotators manually annotated distinct portions of the dataset, and all provided 50 overlapping annotations for measurement of inter-annotator agreement. We hope that providing gold Spanish AMRs to accompany the sentences will enable fuller evaluation of AMR parsers for Spanish and pave the way for additional tasks to be carried out in Spanish that use AMR as an intermediate representation.\footnote{Guidelines and stand-off annotations will be made publicly available on GitHub upon acceptance. The text is available through the Linguistic Data Consortium.}%


Our primary contributions include:
\begin{itemize}
    \item Spanish AMR \textbf{guidelines} which account for a range of relevant linguistic phenomena;
    \item A \textbf{corpus of 586 manually annotated Spanish AMRs} based on our guidelines,
    which will facilitate crosslinguistic comparison and use in evaluation of AMR parsers; 
    \item \textbf{Discussion} on findings from our work and disagreement analysis.
\end{itemize}

\section{Related Work}
\subsection{Abstract Meaning Representation}

\begin{figure}[htb]
    \centering
    \includegraphics[trim={0.5cm 1.5cm 4cm 0cm},clip, scale=0.45]{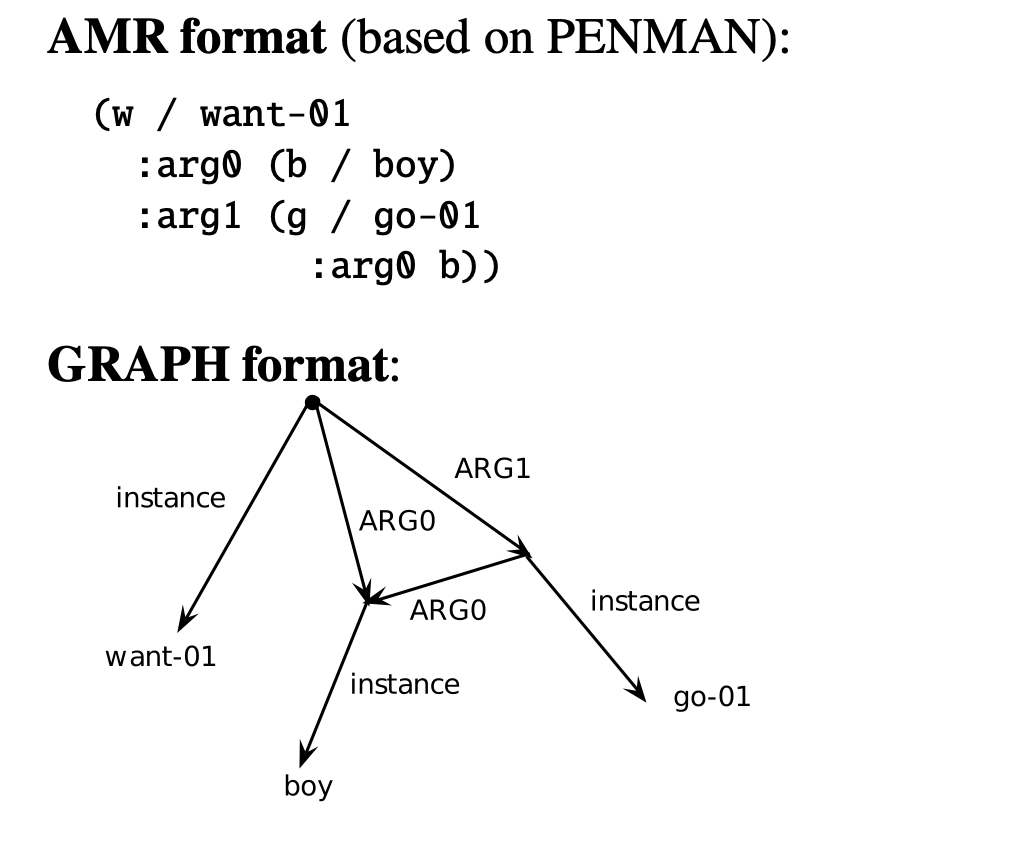}
    \caption{An AMR for the sentence "the boy wants to go," in both PENMAN (text-based) notation as well as in the GRAPH format \citep{banarescu-etal-2013-abstract}.}
    \label{fig:AMR_sample}
\end{figure}

The Abstract Meaning Representation (AMR) framework abstracts away from a language's morphology and syntax, reflecting the core meaning of the sentence or phrase \citep{banarescu-etal-2013-abstract}. The primary goal of the AMR formalism is to capture core semantic elements in a graph structure closer to logic than a syntactic parse; for English this involves removing grammatical details such as number, definiteness, tense, word class (semantic predicates do not distinguish between nouns, verbs, and adjectives), and word order. 

AMR is structured as a rooted, directed graph, where nodes correspond to concepts and labels denote relations between concepts. Labels can be core roles / arguments (marked by \texttt{:argN}), or other attributes such as \texttt{:location} or \texttt{:manner}. \ref{fig:AMR_sample} shows an example annotation of AMR, in both the graph format and the text-based PENMAN notation.
When an entity or event instance participates in multiple roles in the sentence, this is indicated through reentrancy (a shared variable).
Inter-annotator agreement for AMR annotation is measured using Smatch \citep{cai-knight-2013-smatch}, which determines the semantic overlap between the AMRs by comparing concepts and labels.
Here we focus on sentence-level AMR annotations, though additional kinds of cross-sentence relationships within a document have been proposed \citep{msamr}\finalversion{\nss{add DocAMR}}.

\subsection{Cross-lingual Adaptations of AMR}
\label{ssec:adaptations}

While AMR was designed originally for English and not intended to serve as an interlingua \citep{banarescu-etal-2013-abstract}, AMR is particularly well suited to cross-lingual adaptation because AMR strips away morpho-syntactic variation \citep{li-etal-2016-annotating}. However, because AMR was not designed to capture meaning for languages other than English, there are changes to the annotation schema that are required to sufficiently account for language variation and pertinent linguistic phenomena.

Cross-lingual adaptations of AMR have been developed and evaluated for multiple languages, including Czech \citep{hajic-etal-2014-comparing}, Chinese \citep{xue-etal-2014-interlingua,li-etal-2016-annotating}, Spanish \citep{migueles-abraira-etal-2018-annotating}, Vietnamese \citep{linh-nguyen-2019-case}, Portuguese \citep{sobrevilla-cabezudo-pardo-2019-towards,anchieta-pardo-2018-towards}, and Korean \citep{choe-etal-2020-building}. The Uniform Meaning Representation extension of AMR aims to incorporate linguistic diversity into the AMR annotation process, with special consideration for low-resource languages \citep{van_gysel_umr_2021}.

\subsection{Prior Work Adapting AMR to Spanish}

Prior work has proposed and piloted the adaptation of AMR to Spanish \citep{migueles-abraira-etal-2018-annotating}.
This prior approach used English AMR guidelines \citep{AMRGuidelines} as a baseline, piloted annotation for Spanish sentences using the English guidelines, and identified Spanish linguistic phenomena which were not sufficiently captured by the English AMR guidelines. Seven linguistic phenomena were identified and incorporated into the Spanish AMR guidelines: (1)~NP ellipsis, (2)~third person possessive pronouns, (3)~third person clitic pronouns, (4)~varied \emph{se} usage, (5)~ gender, (6)~verbal constructions, and (7)~double negatives.
The Spanish translation of \emph{The Little Prince} was used for annotation during the pilot phase as well as during the actual corpus annotation. However, the Spanish translations were made to be more literal so that they would be more semantically equivalent to the original Spanish translation of the work. 50 Spanish annotations were released in total. The 50~sentences were selected to offer coverage of a range of linguistic phenomena.


\subsection{Limitations of Prior Work on Spanish AMR}
\label{ssec:limitations_prior}
One limitation of the existing Spanish AMR approach is the use of English PropBank \citep{kingsbury-palmer-2002-treebank,palmer_2005_propbank} for sense annotation \citep{migueles-abraira-etal-2018-annotating}. This is an issue because English PropBank senses correspond to the various meanings captured by English words, which do not correspond one-to-one with their Spanish verbs. The implication of the use of English PropBank for role annotation is an undue reliance on English semantics in this approach.

\Citet{migueles-abraira-etal-2018-annotating} chose rolesets from English PropBank instead of AnCora \citep{taule-etal-2008-ancora} because AnCora's coverage is incomplete. Thus, Spanish words were translated to English, and the sense from the English word was attached to the Spanish word \citep{NoeliaMiguelesAbrairaMastersThesis}.
In our work, we compensate for the incompleteness of AnCora by keeping track of any missing senses or verbs and supplementing AnCora senses with our own.

Another limitation of the existing approach to Spanish AMR annotation is the limited amount of change to the English AMR guidelines to incorporate Spanish linguistic phenomena. We describe our design goals, including the linguistic phenomena we prioritize in the Spanish AMR annotation process, in \cref{sec:design_goals}.

Recent work has also assessed the various difference between Spanish and English annotations of the existing Spanish AMR adaptation, classifying the structural divergences and describing the divergences as being due to semantic divergences, syntactic divergences, or annotation choices \citep{wein-schneider-2021-classifying}.

\section{Dataset}

We perform annotations on the ``AMR 2.0 - Four Translations'' dataset, which is released through the Linguistic Data Consortium \citep{damonte_cohen_amr_four_translations}.
This dataset contains gold AMRs for \emph{English} sentences only, alongside translations of those sentences into four languages, including Spanish.
It has become a popular evaluation tool for cross-lingual AMR parsers \citep{blloshmi-etal-2020-xl,procopio-etal-2021-sgl,cai-etal-2021-multilingual-amr}. 
In this work we directly annotate AMRs for the Spanish sentences. Having the English translation is not sufficient for evaluation of Spanish AMRs because of the divergences between English and Spanish AMRs \citep{wein-schneider-2021-classifying} and because of the role that both semantics and syntax play in AMR parsing \citep{damonte_thesis}.
We hope that our release of gold Spanish AMR annotations on the ``AMR 2.0 - Four Translations'' dataset will make for more effective evaluation of cross-lingual parsers.

Additionally, we choose to annotate this dataset with the aim of making our Spanish AMR annotations more general. Though many existing AMR annotated corpora are translations of \emph{The Little Prince} \citep{banarescu-etal-2013-abstract, li-etal-2016-annotating, migueles-abraira-etal-2018-annotating, anchieta-pardo-2018-towards, linh-nguyen-2019-case},
it is a work of fiction with very specific language. We opt to annotate the Four Translations dataset as to more broadly identify linguistic phenomena or concepts which need to be incorporated into the guidelines

The Four Translations dataset contains Italian, Spanish, German, and Mandarin Chinese translations of the English test split sentences from the AMR Annotation Release 2.0 \citep{knight_amr_2.0}. The sentences originate mostly from news sources, including broadcast conversations, newswire and web text. The corpus contains 1,371 Spanish sentences and 5,484 sentences total. 

Of the 1,371 Spanish sentences, we annotate 486. 
There are five documents included in the Four Datasets dataset: Proxy reports from newswire data (Proxy), translated Xinhua newswire data (Xinhua), BOLT discussion forum source data (DFA), DARPA GALE weblog and Wall Street Journal data (Consensus), and BOLT discussion forum MT data (Bolt). For Consensus, Proxy, Bolt, and DFA, we annotate the first 100 sentences of the document. Xinhua is 86 sentences in total, so we annotate all 86 sentences. Consensus is originally 100 sentences, Proxy is originally 823 sentences, Bolt is 133 sentences, and DFA is 229 sentences long.

\section{Annotation Methodology}

\subsection{Annotator Training}
Three undergraduate linguistics students, who are native English speakers with high levels of Spanish proficiency, 
were first trained in English AMR annotation. Subsequently, the annotators were trained in our approach to Spanish AMR annotation, through discussions on our v1.0 Spanish AMR guidelines.
Annotation training for English and Spanish AMR took place on \emph{The Little Prince} corpus, the English and Spanish translations respectively. After piloting English AMR annotation, and then Spanish AMR annotation, the annotators moved on to annotations of the Four Translations dataset.
In order to verify annotator understanding, we completed adjudication on the test sets of English and Spanish annotations.

\subsection{Collected Annotations}

\begin{table}
\begin{center}
\small
\begin{tabular}{ |c|c|c| } 
 \hline
 Document & Sentences & Annotator(s)\\ 
 \hline
 Proxy & 1--50 & 1, 2, 3\\ 
 Proxy & 51--100 & 1\\
 Consensus & 1--100 & 1\\
 Bolt & 1--100 & 2\\
 DFA & 1--40 & 2\\
 DFA & 41--100 & 3\\
 Xinhua & 1--86 (e.o.f.) & 3\\
 \hline
\end{tabular}
\caption{Description of annotations in the corpus, including documents, sentences, and which annotator produced the annotation. Sentence numbers are inclusive.
} 
\label{tab:annos}
\end{center}
\end{table}

To validate our approach to annotation and the reliability of our annotations, we collect annotations from all three annotators for the first 50 sentences from the Proxy document. We are then able to perform inter-annotator agreement analysis on those overlapping annotations using Smatch, presented in \cref{sec:evaluation}.
Other than those 50 Proxy annotations, all other annotations were distributed evenly between each of the three annotators. \Cref{tab:annos} reflects the annotations produced by each (anonymized) annotator. The three annotators produced 200, 190, and 196 annotations each. This results in a total of 586 annotations total, for 486 unique sentences, with Proxy 1--50 being annotated thrice (once by each annotator).
After all annotations for the initial 50 sentences were produced, a final round of corrections were made for any errors in annotation (without changing any divergent judgment calls).

AMR annotation is expensive and time-consuming. Our 586 annotations took more than 200 hours to complete including some test annotations and correction of annotations. This is also a result of the very long and complicated sentences included in the AMR 2.0 - Four Translations dataset, which are especially difficult to annotate due to their genre and length. As a result, we choose to double annotate only a portion of the data (50 sentences) and maximize the number of sentences gold annotated.

\subsection{AnCora}
We use the AnCora-ES lexicon
of verbs for verb sense annotation \citep{taule-etal-2008-ancora}.
The AnCora lexicon is comprised of predicates, accompanied by their argument structures.
Each predicate is also related to one or more semantic classes depending on its senses.

For all verbs or verb senses which did not appear in the AnCora corpus, we kept track of those instances in a table and supplemented the AnCora verb bank with our own. We did not overlap any senses with AnCora so that there was no confusion with the senses being marked in our annotations: if we identified a new verb sense, we assigned the number subsequent to the highest AnCora verb sense.

\subsection{StreamSide Annotation Tool}

Annotations were produced using the Streamside annotation software \citep{choi2021streamside}.
The annotators annotate tokens in the sentence as concepts, and roles and arguments are then defined between these concepts as relations.

While this software allows for annotation fitted to various languages, it is best accustomed to annotation using the English because the relevant PropBank roles \citep{kingsbury-palmer-2002-treebank,palmer_2005_propbank} are automatically populated.
In our case, working on Spanish and using the AnCora rolesets \citep{taule-etal-2008-ancora}, the annotators needed to separately reference the arguments for each concept on the AnCora website.


\subsection{Guidelines Development}
We developed the guidelines by first outlining our approach to key Spanish linguistic phenomena, which we identified as potentially impacting Spanish AMR annotation. Our v1.0 guidelines discuss:
\begin{itemize}
    \item Use of English AMR Roles and Guidelines
    \item Pronoun Drop and NP Ellipsis
    \item Third Person Possessives
    \item Third Person Clitic Pronouns
    \item \w{Se} Usage
    \item Gender
    \item Double Negation
    \item Diminutive and Augmentative Suffixes
    \item \w{Estar} (to be) as a Location
\end{itemize}
These v1.0 guidelines were developed \emph{before} performing any annotation. Since starting annotation, there have been 9 further iterations of the guidelines, which both expand on the items included in v1.0  and incorporate additional items. We discuss the most notable elements of the guidelines in \cref{sec:design_goals}.
After developing the v1.0 guidelines, any further changes required to the guidelines, as identified during the annotation process, were incorporated into the guidelines. All existing annotations were then uniformly altered by their annotators to match the most updated guidelines.

\section{Aims and Guidelines}
\label{sec:design_goals}


Our primary aims with the development of this corpus included the release of a (1)~sizable, (2)~general-purpose Spanish AMR corpus, which can be useful in the evaluation cross-lingual AMR parsers, (3)~which effectively represents Spanish semantics. We set out to meet these goals by (1)~manually annotating 586 AMRs, (2)~annotating the Four Translations dataset, often used for evaluation of cross-lingual AMR parsers, and (3)~developing guidelines which consider a range of linguistic phenomena.

In the subsections that follow, we discuss the key considerations and linguistic phenomena we prioritize in our approach to Spanish AMR annotation.

\subsection{Use of English and Connection to English AMR Guidelines}
\label{ssec:english_use}

Our guidelines are developed in reference to the English AMR Guidelines\footnote{\url{https://github.com/amrisi/amr-guidelines/blob/master/amr.md}}, outlining the differences between our annotation schema of Spanish sentences and the annotation for English AMRs. As has been popularized in other non-English AMR corpora
\citep{linh-nguyen-2019-case,sobrevilla-cabezudo-pardo-2019-towards},
we maintain the role labels and canonical entity type list in English. For example, we use \amr{:ARG0}, \amr{:ARG1}, etc., as well as \amr{:domain}, \amr{:time}, etc., and \amr{person}, \amr{government-organization}, \amr{location}, etc.

\subsection{Verb Senses}
\label{ssec:verb_senses}

We number verb senses according to the AnCora lexicon\footnote{\url{http://clic.ub.edu/corpus/en/ancoraverb_es}}, and supplement these with new senses for out-of-vocabulary lexemes and meanings encountered in our data (\cref{tab:verb_senses}).
Usage examples for these senses are included in the guidelines.

\begin{table}\centering\small
\setlength{\tabcolsep}{2.5pt}
\small
\begin{tabular}{@{}c c c c@{}}
\hline
\textbf{Verb} & \textbf{In AnCora?} & \textbf{Sense} & \textbf{English Translation} \\
 \hline
auditar & no & -01 & to audit \\
disuadir & no & -01 & to dissuade \\
vagar & no & -01 & to wander \\
hervir & no & -01 & to boil \\
desvanecer & yes & -02 & to fade \\
sobrecargar & no & -01 & to overload \\
congestionar & no & -01 & to congest (traffic) \\
incriminar & no & -01 & to incriminate \\
circunvalar & no & -01 & to encircle \\
adular & no & -01 & to flatter \\
salir & yes & -11 & to go out (with someone) \\
entrelazar & no & -01 & to interlace \\
zonificar & no & -01 & to zone \\
embotellar & no & -01 & to bottle up\\
deber & yes & -03 & [modal] to recommend \\
poder & yes & -04 & [modal] to be possible \\
 \hline
\end{tabular}
\caption{Table of verb senses specification for annotation of senses which are not covered by AnCora.}
\label{tab:verb_senses}
\end{table}


\subsection{Modality}
\label{ssec:modality}

The modal verbs \emph{deber} (``must'', ``should'') and \emph{poder} (``might'', ``could'') appear in \cref{tab:verb_senses} in the list of words which appear in AnCora with other senses. Though meanings of \emph{deber} and \emph{poder} do appear in AnCora, we establish additional senses to mark modality. These modals take the same \texttt{:ARG1} structure as do their English modal equivalents---\texttt{recommend-01} and \texttt{possible-01}, respectively. These modals take the verb senses \texttt{deber-03} and \texttt{poder-04}.

\subsection{Gender}
\label{ssec:gender}

In Spanish, all nouns have lexical gender (masculine or feminine), which affects agreement. Nouns relating to humans or animals will also be marked with natural\slash interpretable gender, such as \emph{hermano} (``brother'') versus \emph{hermana} (``sister''). 
Either way, we remove only number information when lemmatizing the word for AMR, 
so \emph{niños} (whether it means ``boys'', or ``boys and girls'') will always be represented with the concept \emph{niño}, and \emph{niñas} (``girls'') with \emph{niña}.


\subsection{Pronoun Drop}
\label{ssec:prodrop}

Spanish belongs to a group of languages that allow pronoun drop (pro-drop), in which certain pronouns can be omitted if they are grammatically or pragmatically inferable from the surrounding linguistic context. Pro-drop in Spanish occurs only with subject pronouns and is permitted only in certain contexts \citep{espanola2010nueva}.\footnote{
Subject drop is viable in Spanish due to inflection of person and number in the verb. Other pro-drop languages
permit the elision of pronouns in other positions. Future work can look at the impact of AMR's abstraction away from morphosyntactic information that allows phenomena such as pro-drop, especially in translation and generation tasks.} 
\Citet{migueles-abraira-etal-2018-annotating} specify a special concept \amr{sinnombre} (``nameless'') for 
implicit references where no antecedent in context is represented in the AMR.
We refine this approach to also encode person and number for these implicit entities:
\begin{itemize}
    \item \amr{first-person-sing-sinnombre}
    \item \amr{first-person-plural-sinnombre}
    \item \amr{second-person-sing-sinnombre}
    \item \amr{second-person-plural-sinnombre}
    \item \amr{third-person-sing-sinnombre}
    \item \amr{third-person-plural-sinnombre}
\end{itemize}

For example, in \emph{No sé que quiero} (``I do not know what I want''), there is an implicit subject \emph{yo} (``I'') 
that is reflected in the verbal agreement.
We therefore specify \texttt{first-person-sing-sinnombre} as the agent. 
We choose to use \amr{first-person-sing-sinnombre} instead of the reentrant \amr{yo} (``I'') as the conditions on the use of overt and dropped pronouns are typically subject to information structure, an important component of sentence meaning.

\smallbreak
\noindent\emph{No sé que quiero.} (``I do not know what I want.'')

\begin{verbatim}
(s / saber-01
    :polarity -
    :ARG0 (f / first-person-sing-sinnombre)
    :ARG1 (h / querer-01
        :ARG0 f ))
\end{verbatim}

If the pronoun is present (i.e. \emph{él, ella, usted,} etc.), the pronoun is should be used in place of a \amr{sinnombre} concept.




 
\subsection{Polite Second Person Addressee}
\label{ssec:polite_usted}

\emph{Usted} (``you'') can reflect either a polite usage of second person, or third person. When \emph{usted} is used as a polite second person pronoun, the polite modifier should be added: \noindent\texttt{:mod-polite +}. This follows the same structure as \noindent\texttt{:polarity -}. 

\smallbreak
\noindent\emph{usted} (``you'')

\begin{verbatim}
(u / usted
    :mod-polite +)
\end{verbatim}


\subsection{Third Person Possessives}
\label{ssec:third_person_possessives}

We treat third person possessives similarly to the English annotation, using the \amr{sinnombre} concepts discussed above. For example, we annotate \emph{su coche} (``his car'') the same way that ``his car'' is structured.
\smallbreak
\noindent his car
\begin{verbatim}
(c / car
    :poss (h / he)) 
\end{verbatim}

\smallbreak
\noindent\emph{su coche} (``his car'')

\begin{verbatim}
(c / coche
    :poss (e / third-person-sing-sinnombre))
\end{verbatim}

The possessive pronoun \emph{su} is ambiguous (``his''/``hers''/``its''), and could be annotated as \amr{third-person-sing-sinnombre} (in the case of ``his''), \amr{second-person-sing-sinnombre} (as in ``yours''), or \amr{third-person-plural-sinnombre} (for ``theirs'').
These labels are only required when the use of \emph{su} as a possessive pronoun is ambiguous. For example, in the case of \emph{Sofía me mostró su auto} (``Sofía showed me her car''), \emph{su} very likely refers to Sofia's. However, in \emph{Sofía copió su tarea} (``Sofía copied their homework''), this likely means that Sofia copied someone else's homework; \emph{su} would refer to some unnamed person, and would thus require the use of \amr{third-person-sing-sinnombre}. Because \emph{su} covers all third person possessives, this distinction requires some interpretation by the annotator based on context and meaning.

\subsection{Third Person Clitic Pronouns}
\label{ssec:third_person_clitics}

Clitics are treated as separate tokens. For example, \emph{mandarlo} (``send it'') has a root of \emph{mandar} (``send'') and an ARG1 of the item being sent: \emph{lo} (``it'').

\begin{verbatim}
(m / mandar-01
    :ARG1 (l / lo))
\end{verbatim}

\subsection{Se Usage}
\label{ssec:se_usage}

\emph{Se} has many uses in Spanish, including: (1)~as a reflexive pronoun, (2)~to denote the passive voice, (3)~as a substitute for the indirect pronoun \emph{le/les}, and (4)~as an impersonal pronoun.


\paragraph{Se as a Reflexive Pronoun.}
Reflexives are represented via reentrancies as in English AMR. Two examples include the use of \emph{se} in \emph{ellos se perjudican} (``they are harmed'') and in \emph{Pablo se ve} (``Pablo sees himself'').

\smallbreak
\noindent\emph{Ellos se perjudican.} (``They harm themselves.'')

\begin{verbatim}
(p / perjudicar-01
    :ARG0 (s / third-person-plural-sinnombre)
    :ARG1 s)
\end{verbatim}
 
\noindent\emph{Pablo se ve.} (``Pablo sees himself.'')

\begin{verbatim}
(v / ver-01
    :ARG0 (p / persona
        :name (n / name
            :op1 "Pablo"))
    :ARG1 p)
\end{verbatim}
 
\paragraph{Se as a Passive Marker.}

When se reflects a passive voice for an omitted concept, we use the \texttt{:prep-by} role label with \emph{se}.

\smallbreak
\noindent\emph{Se venden casas rurales.} (``Rural houses for sale.'')

\begin{verbatim}
(v / vender-01
    :ARG0 (s / se)
    :ARG1 (c / casa
        :mod (r / rural)))
\end{verbatim}
 
\paragraph{Se as a Substitute for ``Le'' or ``Les.''}

When the indirect object pronoun \emph{le} or \emph{les} is  directly followed by a pronoun which starts with an \emph{l}, \emph{le/les} is substituted by \emph{se}.
With \emph{le} acting as an unnamed entity, follow the guidelines per \cref{ssec:prodrop}. If it is referring to a named (specific) entity, we refer back to that entity in the AMR.

\paragraph{Se as an Impersonal Pronoun.}

\emph{Se} used to mean ``one'' is annotated with the concept \texttt{se-impersonal}:

\smallbreak
\noindent\emph{No se debe beber.} (``One should not drink.'')

\begin{verbatim}
(d / deber-01
    :polarity -
    :ARG0 (b / beber-01)
    :ARG1 (s / se-impersonal))
\end{verbatim}

\subsection{Double Negation}
\label{ssec:negation}

In Spanish, negation can be indicated by either single or double negatives, with double negatives sometimes providing emphasis.
We annotate both single and double negation with the use of one polarity marker.

\smallbreak
\noindent\emph{No hay ninguna persona.} (``There is nobody.'')

\begin{verbatim}
(h / haber-01
    :polarity -
    :ARG0 (p / persona))
\end{verbatim}

 
\subsection{Suffixes}
\label{ssec:suffixes}

Derivational suffixes such as diminutives should be represented as modifier concepts. For example, \emph{poquito} (``very little'') would be annotated with \emph{poco} (``little'') being modified by \emph{muy} (``very'').
 
\begin{verbatim}
(p / poco
    :mod (m / muy))
\end{verbatim}
 
Another example would be \emph{hombrecito} (``little man''), for which would \emph{hombre} (``man'') receive the diminutive modifier of \emph{pequeño} (``little'').

\begin{verbatim}
(h / hombre
    :mod (p / pequeño))
\end{verbatim}

\subsection{Words that Change Meaning When Singular Or Plural}
\label{ssec:plural_singular}

In Spanish AMR as in English AMR, we annotate the concept as the singular of the entity even if it is plural. However, rarely in Spanish a word changes meaning if it is plural instead of singular. In this case we use the plural form of the word.
Additionally, we distinguish \emph{algún} from \emph{algunos}, for the case in which \emph{algún} means ``any'' and \emph{algunos} means ``some.''
Similarly, we distinguish \emph{otros} (``others'') as a plural noun to mean a distinct group of ``others,'' and preserve the plural \emph{otros} instead of making it singular as \emph{otro} (``other'').

\subsection{Commas and Decimals}
\label{ssec:commas_decimals}

When using numerical decimals as they appear in the Spanish sentence (with a comma rather than a period, as in American English), we preserve the Spanish use and indicate decimals using a comma.
For large numbers in the thousands or millions, for example 1 million, these are expressed without any place markers, as is also done in the English AMR Guidelines.

\subsection{Comparison with Previous Work}


The most notable difference between our approach and that of \citet{migueles-abraira-etal-2018-annotating} is that theirs uses Spanish labels while ours uses English labels, which is a choice we discuss in \cref{ssec:limitations_prior}. Additional differences are largely due to our choice to break down the unnamed category of dropped entities into subcategories based on the type of noun phrase or pronoun. For NP Ellipses (\cref{ssec:prodrop}) and third person possessives (\cref{ssec:third_person_possessives}), we use the 6 tags outlined, which specify person and number. \Citet{migueles-abraira-etal-2018-annotating} uses a standardized \amr{ente} (``being'') concept with \amr{sinnombre} (``nameless'') argument for NP ellipses and a \amr{sinespecificar} (``unspecified'') argument for third person possessives.
In comparison to our annotation in \cref{ssec:third_person_possessives} for \emph{su coche} (``his car''), the annotation in \citeposs{migueles-abraira-etal-2018-annotating} corpus would be: 
\begin{verbatim}
(c / coche
    :posee (e / ente
        :sinespecificar (s / su)))
\end{verbatim}
We make the choice to further subdivide these categories to be able to capture relevant semantic information.

Our approach as well as that of \citet{migueles-abraira-etal-2018-annotating} represents clitics as if they were separated from the stem. We also both approach \emph{se} as a reflexive pronoun in the same way via reentrancy. However, the approach of previous work omits \emph{se} when it is used in the impersonal or passive voice, which we include via the \texttt{se-impersonal} concept and \texttt{prep-by} label, respectively (\cref{ssec:se_usage}). 
We also address the issues of \emph{se} as a substitute for \emph{le} or \emph{les} (\cref{ssec:se_usage}), modality (\cref{ssec:modality}), gender (\cref{ssec:modality}), polite use of \emph{usted} (``you'') (\cref{ssec:polite_usted}), double negation 
(\cref{ssec:negation}), diminutive and augmentative suffixes (\cref{ssec:suffixes}), meaning change in the singular versus plural (\cref{ssec:plural_singular}), and commas/decimals (\cref{ssec:commas_decimals}).

\section{Evaluation}
\label{sec:evaluation}

\subsection{Inter-Annotator Agreement}

\Cref{tab:iaa} shows the inter-annotator agreement (IAA) scores for each pair of annotators on the 50 triple-annotated Proxy sentences. The IAA scores were calculated by averaging the Smatch scores across the 50 sentence pairs for the annotators. 
The Smatch \citep{cai-knight-2013-smatch} algorithm calculates the amount of overlap between the AMR graphs to determine similarity. 

The average IAA scores ranged from 0.83--0.89, a very promising range for AMR annotation agreement. Comparable work achieved Smatch inter-annotator agreement scores of 0.79 \citep{choe-etal-2020-building}, 0.72 \citep{sobrevilla-cabezudo-pardo-2019-towards}, and 0.83 \citep{li-etal-2016-annotating}.
Other work on cross-lingual AMR adaptations did not report IAA\slash Smatch scores, often because it was only annotated by one annotator.
In the subsections that follow we discuss common disagreements and annotator mistakes.

\begin{table}[h!]
\begin{center}
\small
\begin{tabular}{ |c|c|c| } 
 \hline
 Ann.~1 \& Ann.~2 & Ann.~1 \& Ann.~3 & Ann.~2 \& Ann.~3 \\ 
 \hline
 0.89 & 0.86 & 0.83 \\ 
 \hline
\end{tabular}
\caption{Average inter-annotator agreement scores (via Smatch) for each pair of our three annotators on the first 50 sentences of the Proxy document.} 
\label{tab:iaa}
\end{center}
\end{table}

\subsection{Disagreement Analysis}

Disagreements, which we define as any discrepancy that neither violates AMR guidelines nor deviates from the sentence’s meaning, were common among all three annotators.
The majority of disagreements are caused by differences in interpretation.

\paragraph{Entity versus Event Annotation} For example, AMR takes a verb-centric approach to annotation. While verbs are typically annotated as events and nouns are annotated as entities (concepts without a number),
when nouns or phrases have verbal counterparts, this can cause differences among annotators. In the examples below, \emph{propuesta} (``proposal'') is annotated both as a noun and as a verb. This then has the effect of altering the role of \emph{funcionario} (``official'').

\smallbreak
\noindent\emph{Ignoran la propuesta del funcionario.} (``They ignore the official's proposal.'')

\textbf{Desired Annotation:}
\begin{verbatim}
(c0 / ignorar-01
    :ARG0 (c1 / third-person-plural-sinnombre)
    :ARG1 (c2 / thing
        :ARG1-of (c3 / proponer-01
            :ARG0 (c4 / funcionario))))
\end{verbatim}

\textbf{Possible Annotation 2:}
\begin{verbatim}
(c0 / ignorar-01
    :ARG0 (c1 / third-person-plural-sinnombre)
    :ARG1 (c2 / propuesta
        :poss (c3 / funcionario)))
\end{verbatim}


The desired annotation captures both the act of proposing and the content of the proposal (thing which was proposed), whereas the other options focus on one aspect or the other.

\paragraph{Verb Sense Labels} Additional disagreements arose from differing interpretations of verb senses in AnCora rolesets. Verb senses account for nuance in a verb’s meaning depending on context. Sometimes annotators chose different rolesets when the meaning difference between senses was subtle. One notable example is the verb \emph{reconocer} (``to recognize / acknowledge''). \texttt{Reconocer-01} refers to recognizing something as official or true, as in \emph{reconocer el estado} (``to recognize the state''). Alternatively, \texttt{reconocer-02} maintains that meaning, but often precedes an independent clause, as in \emph{reconocen que gané} (``they acknowledge that I won''). This subtle distinction in meaning may not be as clear in the data, and thus may not be annotated consistently among annotators.



\paragraph{Non-Core Role Overlap} Finally, annotators had difficulty consistently choosing the same non-core role (\texttt{:poss}, \texttt{:mod}, etc.)\ when the roles could overlap in meaning. For example, for the Spanish \emph{la carta del hombre} (``the man’s letter''), this could be annotated differently depending on interpretation of the man’s relationship to the letter. An emphasis on the man’s ownership of the letter elicits the \texttt{:poss} role, whereas emphasizing the letter’s creation by the man elicits the \texttt{:source} role. 



\subsection{Common Mistakes}

Both mistakes and disagreements are unique to each annotator and result in decreased inter-annotator agreement scores. 


In order to preserve continuity with the English AMR guidelines, we use the English role labels and the canonical list of entity types. Early confusion with when to use Spanish tokens versus English labels led to errors. Examples of these errors include annotating gender and translating English non-core roles or named-entity concepts (person, country, city, etc.) into Spanish.

\smallbreak
\noindent\emph{Estados Unidos} (United States)

\textbf{Correct Annotation:}
\begin{verbatim}
(c0 / country
     :name (c1 / name
	          :op1 "Estados"
	          :op2 "Unidos"))
\end{verbatim}

\textbf{Incorrect Annotation:}
\begin{verbatim}
(c0 / país
     :nombre (c1 / nombre
        :op1 "Estados"
        :op2 "Unidos"))
\end{verbatim}




Other miscellaneous errors included spelling errors, adding periods to the concept corresponding to the last token in the sentence, and incorrect verb senses and arguments.

\section{Discussion}


\subsection{Gender and Number Marking}
The construction of Spanish interpretable/natural gender and its relationship to morphosyntax are open questions \citep{donatelli2019morphosemantics}. 
In our annotation schema, we opted for simplicity, choosing not to explicitly annotate gender, but to leave any gender-bearing morphology as is in the concept. 
Like in English AMR, number inflection is removed 
unless that would alter the meaning of the stem (\cref{ssec:plural_singular}).
The possibility of encoding number and gender more explicitly is left to future work.


\subsection{Idiomatic Expressions}

Idiomatic expressions are difficult to annotate with AMR. As is the case for English, Spanish has numerous idiomatic expressions, phrases that have a meaning different to that of individual words in the phrase. Idiomatic expressions are annotated on a case-by-case basis. In the corpus, the majority of idiomatic expressions are either condensed into one concept (\emph{por supuesto}, ``of course,'' becomes \texttt{por-supuesto}), or we must use a similar, pre-existing verb to convey the expression’s meaning, such as \emph{tener prisa} (``to be in a rush'').

\smallbreak
\noindent\emph{Por supuesto que la amo.} (``Of course I love her.'')

\begin{verbatim}
(c0 / amar-01
     :ARG0 (c1 / first-person-sing-sinnombre)
     :ARG1 (c2 / la)
     :manner (c3 / por-supuesto))
\end{verbatim}


\subsection{Limitations with AnCora}
\label{ssec:ancora_limitations}

AnCora's predicate lexicon only includes verbs, unlike English PropBank \citep{palmer_2005_propbank}, which has been extended beyond verbs to include noun, adjective, and complex predicates \citep{bonial-14}. AnCora notably lacks adjective frames and numerous idiomatic\slash phrasal verbs. This posed a challenge when annotating many adjectives and (often more colloquial) verb phrases. For idiomatic verb usage, and it is easy in these cases for annotators to default using the structure of the equivalent English idiomatic structure, and substitute Spanish tokens into the English structure.
Some AnCora rolesets were missing important core roles. Expanding AnCora or similar Spanish propbank efforts would enhance any AMR annotations relying on it.

\subsection{Prepositions}

Prepositions can also be awkward to annotate, if the preposition does not clearly correspond to a non-core role.
Most prepositions we encounter are annotated such that the preposition is captured by a non-core role, such as \texttt{:location} or \texttt{:time}. However, some prepositions describing a specific spatial or temporal relation are included in the AMR itself. In these cases, the preposition is annotated as a concept, and the object of the preposition is given the \texttt{:op1} role. 

\smallbreak
\noindent\emph{El gato duerme bajo la cama.} (``The cat sleeps beneath the bed.'')
\begin{verbatim}
(c0 / dormir-01
     :ARG0 (c1 / gato)
     :location (c2 / bajo
        :op1 (c3 / cama)))
\end{verbatim}

One unexpected challenge that the annotators encountered related to mapping Spanish prepositions with their English AMR core and non-core labels. 
For example, the preposition \emph{con} (``with'') commonly aligns with roles such as \texttt{:accompanier} in the case of \emph{Él baila con la chica} (``he dances with the girl'') or \texttt{:instrument}, such as in \emph{cortar con un cuchillo} (``cut with a knife''). This alignment is not always straightforward, however.
The Spanish preposition \emph{a} (to), for example, can sometimes be denoted by either \texttt{:manner}, in the case of \emph{lo hice a mi manera} (``I did it my way''), or by \texttt{:destination}, in the case of \emph{nosotros fuimos a la ciudad} (``we went to town''). Because we implicitly map the Spanish prepositions to English role labels, this can sometimes be difficult because the mapping between English preposition senses and Spanish preposition senses is not one-to-one.

\subsection{Mood}

Spanish exhibits three grammatical moods: indicative, imperative, and subjunctive. English AMR assumes all sentences to be in indicative mood unless otherwise marked. There are two categories for additional moods: imperatives are marked with \amr{:mode~imperative} and expressive utterances with \amr{:mode~expressive}. As this is a very rudimentary treatment of the semantics of mood, we choose not to adapt it for Spanish AMR. Future work will look more closely at how to integrate the subjunctive mood into Spanish AMR at both the verbal and sentential level. We believe this project merits a larger investigation of the semantics of mood across languages, as categories and criteria are complex both syntactically and semantically.

\section{Conclusion}

We have presented an approach to annotation of Abstract Meaning Representation for Spanish, which considers a range of details and linguistic phenomena relevant to the annotation schema. Following this approach (and in conjunction with the fleshing out of the approach), we  have developed a general-purpose,  500+ annotation corpus of Spanish AMRs, annotated by three trained annotators. Our approach achieves reasonable inter-annotator agreement (0.83--0.89 IAA via Smatch), and we provide an analysis of the disagreements and common errors. Our annotations are on the ``AMR 2.0 - Four Translations'' dataset \citep{damonte_cohen_amr_four_translations}, and we hope that our gold AMRs for the dataset will be useful in cross-lingual AMR parsing.

In this work, we aim to balance continuity with English and prior cross-lingual AMR guidelines with prioritizing the semantic meaning of the Spanish sentences.
\finalversion{Many AMR adaptations use English annotation as a baseline and then make a set of adaptations for linguistic phenomena specific to the other language (or which do not appear in English). The adaptations varied in degree of reliance on English annotations and resources, ranging from simply working with the English AMR guidelines as a baseline and extending them, to using English PropBank for sense annotation \citep{migueles-abraira-etal-2018-annotating} or aligning English and Portuguese sentences and translating English annotations to their cross-lingual framesets \citep{sanches-duran-aluisio-2015-automatic}.}
The effect of using English as a baseline is unspecified but apparent. In order to decenter English as the focus of cross-lingual AMR annotation, we (1)~use in-language rolesets for sense annotation, though there are limitations associated with AnCora as well (\cref{ssec:ancora_limitations}), and (2)~establish guidelines which consider a range of linguistic phenomena \emph{before} doing any annotation. Still, our approach relies on English AMR labels and uses English guidelines as a foundation for our approach.





\section*{Acknowledgments}

We thank reviewers for helpful feedback and the coordinators of the Georgetown University RULE (Research-based Undergraduate Linguistics Experience) program.
This work is supported by a Clare Boothe Luce Scholarship.

\section{Bibliographical References}\label{reference}

\bibliographystyle{lrecnat}
\bibliography{lrec2022-example}



\end{document}